\documentclass[letterpaper]{article}
\usepackage{alifeconf}

\title{Walk the Robot:
\\ Exploring Soft Robotic Morphological Communication driven by Spiking Neural Networks}
\author{\\Matthew Meek$^{1}$, Guy Tallent$^{1}$, Thomas Breimer $^{1}$, James Gaskell $^{1}$, Abhay Kashyap$^{1}$, Atharv Tekurkar $^{1}$,\\ Jonathan Fischman$^{1}$, Luodi Wang$^{1}$, Viet-Dung Nguyen$^{1}$ \and John Rieffel $^1$ \\
\mbox{}\\
$^1$Computer Science Department, Union College, Schenectady NY, USA\\
rieffelj@union.edu}

\begin{document}
\maketitle

\begin{abstract}
Recently, researchers have explored control methods that embrace nonlinear dynamic coupling instead of suppressing it. Such designs leverage dynamical coupling for communication between different parts of the robot. Morphological communication refers to when those dynamics can be used as an emergent “data bus” to facilitate coordination among independent controller modules within the same robot.

Previous research with tensegrity-based robot designs has shown that evolutionary learning models that evolve spiking neural networks (SNN) as robot control mechanisms are effective for controlling non-rigid robots \cite{morph2010}. Our own research explores the emergence of morphological communication in an SNN-based simulated soft robot in the EvoGym environment.

\end{abstract}

\section{Introduction}

Traditional engineering techniques strive to avoid non-linear dynamic coupling between separate components. Generally, non-linear dynamic coupling has been considered too complex and unwieldy to manage effectively. In fact, much energy has been put into suppressing this coupling in a variety of applications with intelligent design and dampening techniques. This approach is well-intentioned, as if left uncontrolled, dynamic coupling can lead to unanticipated behavior or even large-scale failure. 

Despite this potential for catastrophe, many biological systems embrace and exploit dynamic coupling. This phenomenon is seen on a variety of scales, from the structure of proteins \cite{ingber1998architecture} to the structure of cellular cytoskeleton \cite{valero2007tendon}. If dynamic coupling can be found in biology, it posits that such a system may have valid application in the realm of robotics as well.

In 2009, Rieffel \cite{rieffel2010morphological} demonstrated that a highly complex mechanical system is capable of exploiting dynamic coupling as an emergent 'data bus' for communication between disjoint strut modules. By turning off the output of one module network and showing a change in a distal network, it was demonstrated that morphological communication was occurring between separate robot components. The robotic subject of this research were tensengrity robots, which feature rigid elements such as metal struts connected by tensile elements such as strings. 

In this paper, we present an alternative paradigm for the demonstration of morphological communication in robots using simulated robots with both rigid and soft components. We believe that such a design more closely mirrors the reality of marco elements in biology such as a human hand. Research in biology has demonstrated that the human hand utilizes morphological communication for lower-end coordination, allowing neural pathways to specialize for high level tasks \cite{valero2007tendon}. Our robotic simulator \cite{EvoGym} enables support for robots with both rigid and soft components, making it an excellent testbed for experiments in soft robotics. 

Thus, we present a framework where soft robot designs can be tested, and through the use of an evolutionary algorithm, \cite{nomura2024cmaessimplepractical}, learn to produce a walking gait.
\section{Background}
\subsection{Evolutionary Robots}
Evolutionary robotics has been the fascination of the robotics field for quite a few years. There exist non-adaptive and non-autonomous robots, such as those in factories, and adaptive yet non-autonomous robots, such as drones. The exciting frontier of evolutionary robotics concerns robots that are both adaptive and autonomous - robots that learn and adapt by themselves. \cite{evobots-background}
\\\\
Advances in evolutionary robotics have led to the development of a variety of non-traditional body plans for robots. Robots with non-traditional body plans, such as tensegrity robots, modular robots, and swarm bots, are notoriously difficult to control with conventional control schemes \cite{evobots-background}. Tensegrity robots, for example, are collections of struts, strings, and servo motors held in self-stressed equilibrium. They move with a quasi-static gait that exploits the nonlinear dynamic coupling between the components. Such gaits are exceedingly challenging to develop and optimize by hand. Previous research with tensegrity robots successfully used evolutionary methods to develop gaits instead. \cite{morph2010}.
\\\\
Soft robots are another kind of robot that uses a non-traditional body plan. Soft robots are robots constructed from soft, compliant materials. Their non-rigid nature makes them difficult to control like other types of robots with non-tradition body plans. Evolutionary methods for controlling soft robots have already shown some promise \cite{evobots-background}.

\subsection{EvoGym}
EvoGym is a 2-dimensional soft robotics simulator created with the goal in mind to co-optimize robot design and control. The simulator provides a platform that covers a wide range of tasks that include walking, object manipulation, climbing, locomotion, shape change, flipping, jumping, and balancing. In addition to this, robots can interact with various types of terrains. The robot design interface of the simulator allows users to select different voxels types, with some being fixed and soft voxels, as well as horizontal and vertical actuators. EvoGym also includes deep reinforcement techniques that make the simulator an optimal tool for understanding how robot designs evolve when it comes to completing tasks.

\subsection{Spiking Neural Networks}

Developing locomotion in robots with non-linear dynamical coupling is complex and goes far beyond a simple linear relationship between input telemetry and output. Non-linear relationships are well modeled by Neural Networks which possess a number of layers that operate sequentially on the inputs to determine an output, and perform much better than other methods within the machine learning class on problems of this type.

A traditional neural network is a computational model inspired by the structure of a biological brain. It consists of an input layer, hidden layers, and an output layer, where each layer of the network is made up of neuron nodes. Each neuron receives input signals, processes them using its associated weights and biases, and outputs information to the next layer \cite{vreeken2003spiking}. The connections between neurons, called synapses, are modeled by adjustable scalar weights. A bias is a threshold that allows neurons to activate even when the weighted sum of their inputs is not sufficient. Weights and biases are randomized, to begin with, and are adjusted over time. Most Neural Networks employ a training phase and require tagged data to "discover" patterns within data which can be exploited to predict correct values from inputs. In our case we do not know the optima of the fitness landscape so must determine another way to discover weights.

Further, unlike traditional artificial neural networks, which output continuous values, Spiking Neural Networks (SNNs) operate with discrete events called "spikes” \cite{maass2001pulsed}. When a neuron receives sufficient input, it generates an action potential spike. SNNs encode information on the precise timing of individual spikes that mimic real neuronal activity.

Soft robots, with their flexible and deformable bodies, rely on morphological communication, a process that utilizes physical forces and motion as a source of information to 'outsource' some portion of control of a body to its complex anatomical structures. It leverages dynamic mechanical coupling between various parts of a system and relies on precise timing and synchronization \cite{morph2010}. An SNN is particularly well suited for this type of control, as it can synchronize with the robot’s natural mechanical responses. This will make the evolving movement strategies to be both adaptive and efficient.
    
This study explores how an evolved SNN can generate rhythmic locomotion patterns for a soft robot while demonstrating how neural dynamics interact with physical morphology to achieve effective, biologically inspired motion.

\subsection{CMAES}
A pivotal component of this research is the selection and development of effective robot control schemes. Ideally, we would like for our control scheme to be closed-loop and decentralized for the facilitation of morphological communication between components. As discussed, \cite{rieffel2010morphological} presents a decentralized framework for robotic control using multiple SNNs. Thus, we adopt that control scheme for the robots in this study. 

However, an issue arises in the selection of weights and biases for the SNN's. In the case of our principal robot design, its SNN controller requires 72 separate weights and biases. With dynamic coupling as a design feature, it is beyond the capabilities of humans to intelligently select these values. Rather, we use an evolutionary algorithm to pick weights and biases for us.

Evolution in its basic form can be defined as the process
by which individuals compete for limited resources, and
then pass on their genes with some correlation to how fit
they are. Since the low-fit individuals are less likely to
pass on their genes, the fitness of the population naturally
trends upward, at least in theory. This biological process
served as inspiration for a class of Evolutionary Algorithms,
which aim to iteratively discover solutions through random
mutation and interaction with the environment, and without
any experimenter-provided intuition about the problem space
they are intended to search \cite{brooks1991intelligence}. Using this technique, novel
solutions can be discovered by the algorithm, which are often
too complex for human understanding.

For our robots to develop a walking gait, we must use an optimization algorithm to develop suitable SNN weights and biases. For this purpose we selected a covariance matrix adaptation evolution strategy (CMA-ES), which is a stochastic, black-box optimization method for non-convex continuous optimization. As with any evolution algorithm, new candidate solutions are sampled from a multivariate normal distribution– this is an individual’s genome. Pairwise relationships between variables are represented using a covariance matrix, and the adaptation of the covariance matrix allows for the system to generate a second-order model of the fitness function. As a consequence, CMA-ES performs a principal component analysis of successive evolutionary steps, which causes a “drift” towards fitter solutions. For our purposes, we use cmaes, a python-based CMA-ES implementation \cite{nomura2024cmaessimplepractical}.

\section{Methods}

\subsection{Morphology}

The morphology of the robot is crucial. In our work, we prioritized designing a robot that would be able to effectively meet our desired gait.  This process involved testing and sampling different robot structures with various voxel arrangements in order to determine the ideal configuration. Another critical aspect of the morphology was understanding the anatomy of a voxel, as this allowed us to track the positions of individual voxels.
\\

Desirable Gait:

\noindent The ideal gait we were looking for was a walking or running gait, with the goal of getting the robot to walk as far as possible in the positive x direction.
\\
\begin{figure}[ht]
    \centering
    \includegraphics[width=0.4\textwidth]{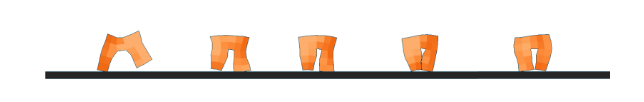}
    \caption{Desirable Gait \cite{EvoGym}}
    \label{fig:gait}
\end{figure}

Voxel Types:

\noindent Robots are stored in json files. Integers 1-5 encode the type of voxel at each point in the robot. 
The types of voxels that make up a robot are:
0: Empty, 1: Rigid, 2: Soft, 3: Horizontal Actuator, 4: Vertical Actuator, 5: Fixed. 
Rigid voxels are non-actuated black voxels that are resistant to expansion and contraction. Soft voxels are non-actuated grey voxels that expand and contract freely, but are not directly controlled. Horizontal actuators are actuated orange voxels that expand and contract horizontally. Vertical actuators are actuated blue voxels that expand and contract vertically. Fixed voxels are non-actuated black voxels that cannot expand, contract, or move \cite{EvoGym}. We used them to construct the platform the robot walks on. 
\\

\begin{figure}[ht]
    \centering
    \includegraphics[width=0.4\textwidth]{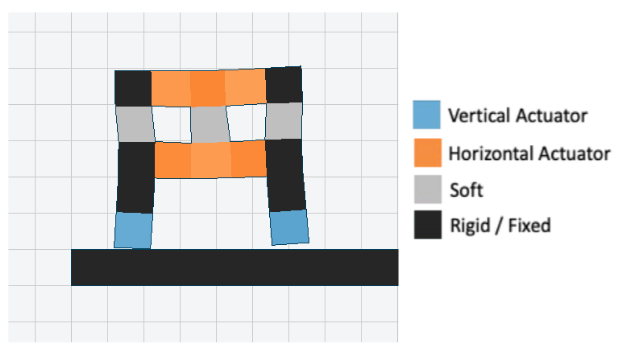}
        \caption{Ubot.json in the simulation. The actuators get lighter when they expand, darker when they contract. \cite{EvoGym}}
    \label{fig:json}
\end{figure}

Ubot:

\noindent In our preliminary experimentation with Evogym’s built-in evolution methods and observation of other robots, we found that a robot with two legs, horizontal actuators to shift them back and forth, and vertical actuators to give height to steps produced desirable results. This style of robot consistently developed a natural, animal-like running gait. We believed a gait like this was a realistic expectation of the type of gait a real-world robot could effectively utilize. This gait became our target when implementing our own evolution and control methods. \ref{fig:json}
\\

\begin{figure}[ht]
    \centering
    \includegraphics[width=0.4\textwidth]{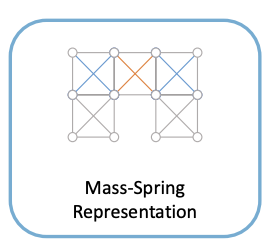}
    \caption{Mass Spring Representation. \cite{EvoGym}}
    \label{fig:massspring}
\end{figure}

Voxel Anatomy, Point Masses:

\noindent Voxels are composed of point-masses and springs. Point-masses define the outer corners of a voxel and are used to track the robot’s position and give weight \cite{EvoGym}.	
We have target lengths for each actuator that range from 0.6 to 1.6. The springs expand or contract to make the actuator match that range. The actuators do not fully expand or contract in a single step. 
\\

\begin{figure}[ht]
    \centering
    \includegraphics[width=0.4\textwidth]{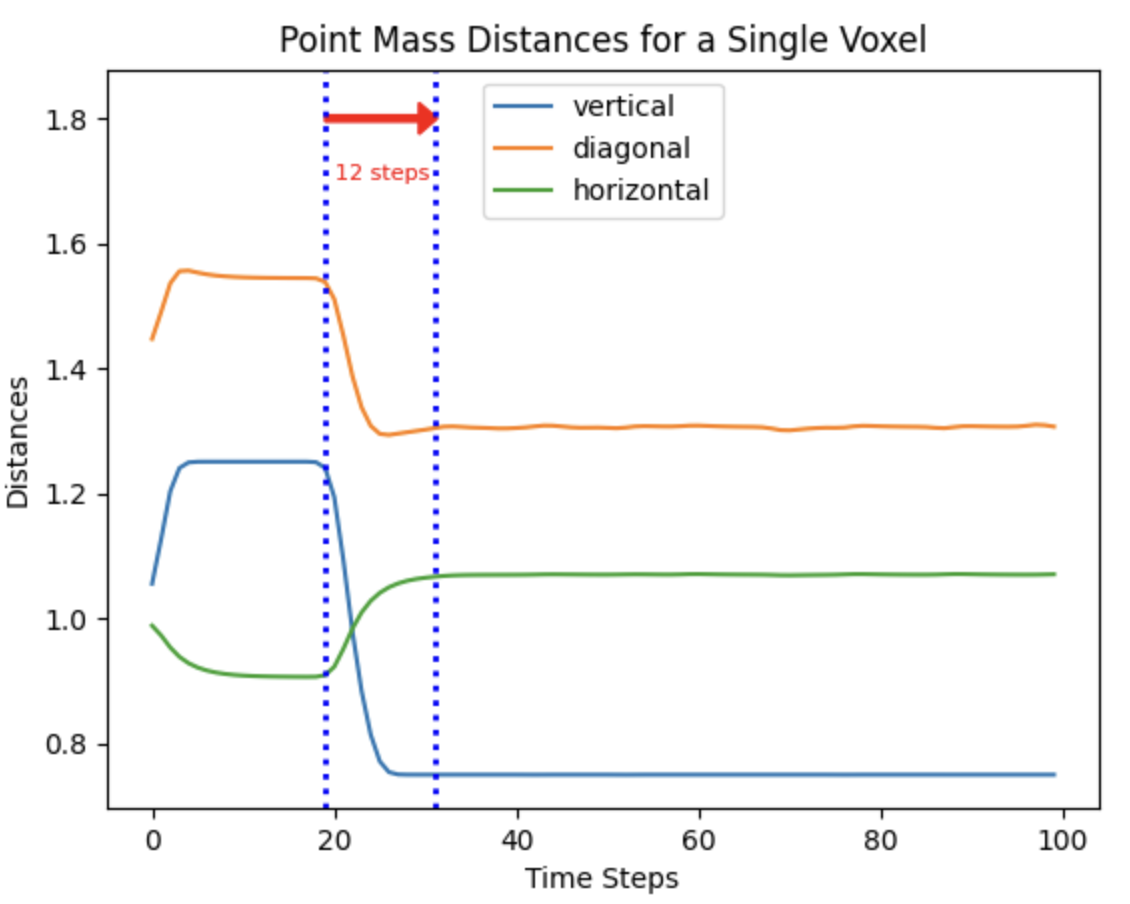}
    \caption{{Point mass distances for a single voxel with manually altered target lengths}}
    \label{fig:12_steps}
\end{figure}

Voxel Behavior, Target Lengths:

\noindent The behavior of each actuator voxel is dictated by a target length and the physics of the simulated springs. This means it takes a number of time steps for the voxel to fully expand or contract to meet its target, which has important implications for our control method. Should the control be overly active, the expansions and contractions will cancel out without the robot being propelled forward. The distance profile for point masses in a single voxel and the time it takes to transition between fully contracted (0.6) and fully expanded (1.6) is shown in Figure \ref{fig:12_steps}.

It takes approximately 12 time steps for the distances between voxels to stabilize; changing the target distances during this transition may have a negative impact on the robot's gait.
\\

Generations/Iterations flow:

\noindent Each run of our program follows this flow chart on the Evogym level. Each generation runs multiple simulations of the robot. Each iteration in a generation is a single step of each Evogym simulation. At each step, we get the robots’ telemetry. The telemetry is the distance from the center of mass of each actuator to the outer four corners of the robot. The telemetry from each actuator is sent to the corresponding SNNs, which return a collection of target lengths for the actuators. The target lengths are applied, and then the simulation progresses a step. We also update the reward function by adding how far it has moved to the existing recorded distance. After each generation, we change the weights in the robots’ SNNs based on how that generation performed. Each robot in the next generation has a set of weights derived from the best robot of the previous generation. At the end, we retrieve a spreadsheet with the best score of each generation and the associated weights.
\\
\begin{figure}[ht]
    \centering
    \includegraphics[width=0.4\textwidth]{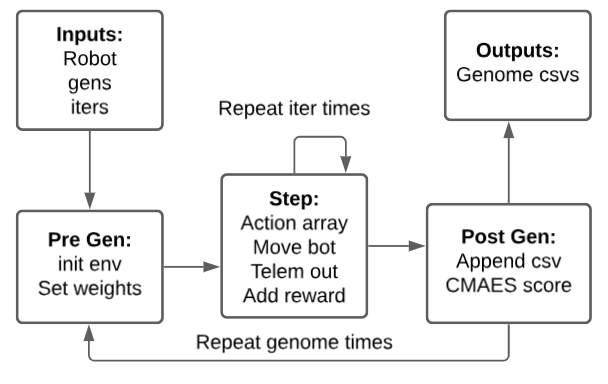}
    \caption{Simulation flow chart}
    \label{fig:gensiters}
\end{figure}



\subsection{Experimental Framework}

\begin{figure}[ht]
  \includegraphics[width=\columnwidth]{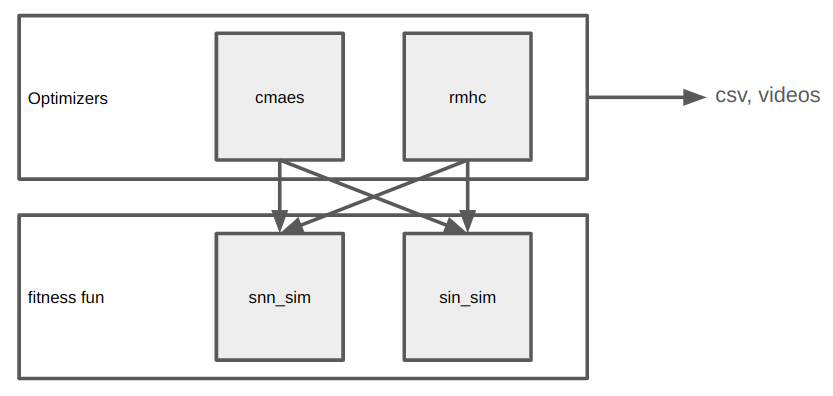}
  \caption{Optimizers and fitness functions need not be intertwined}
  \label{fig:interchangable}
\end{figure}

Our framework is separated into three distinct yet interconnected parts- a CMA-ES module which is responsible for running our evolutionary algorithm, a SNN-Sim package which manages the Evogym lifecycle, and a SNN package which instantiates the SNN controller for each robot. A key feature of this framework is the interchangeability of the separate components. By switching out the simulation package we use, we can easily test robots with different control schemes, such as an open-loop controller with sin-wave actuator activation patterns (Figure \ref{fig:interchangable}). In addition, we may switch out the CMA-ES package and replace it with another optimization algorithm, such as a more traditional evolutionary algorithm or a random mutation hill climber (RMHC).

\begin{figure*}[ht]
  \includegraphics[width=\textwidth]{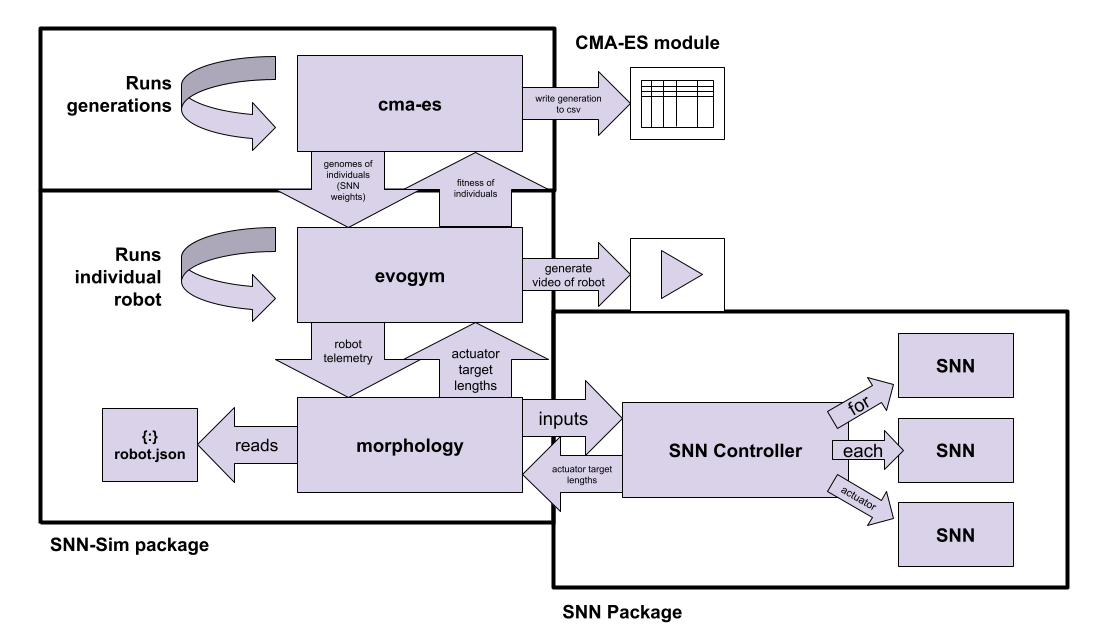}
  \caption{The complete experimental framework}
  \label{fig:complete_framework}
\end{figure*}

\section{CMA-ES}

We begin with the driver of any experimental run, the CMA-ES module. This component is responsible for the generation of new robot genomes, sending them to a fitness function to obtain a fitness for that individual, and continuing to generate new, ideally fitter, genomes. As hyperparameters, CMA-ES takes two inputs- a mean genome and a sigma which CMA-ES uses to generate initial genomes. For our initial experiments, we used a generation size of 12, meaning that 12 different genomes were sampled using a multivariate normal distribution. This normal distribution centers on a mean genome vector with 72 zeros, each representing a mean weight or bias, and with a sigma of 1.

For each genome, we run the fitness function and receive a fitness value for the individual. This value is paired with its respective genome, and we update CMA-ES using this information. Using this information, CMA-ES updates pairwise relationships between genome variables in its covariance matrix, and the adaptation of the covariance matrix allows for the system to generate a second-order model of the fitness function. As a consequence, CMA-ES essentially performs a principal component analysis of successive evolutionary steps, which causes a “drift” towards fitter solutions.

\begin{figure}
    \centering
    \includegraphics[width=1.0\columnwidth]{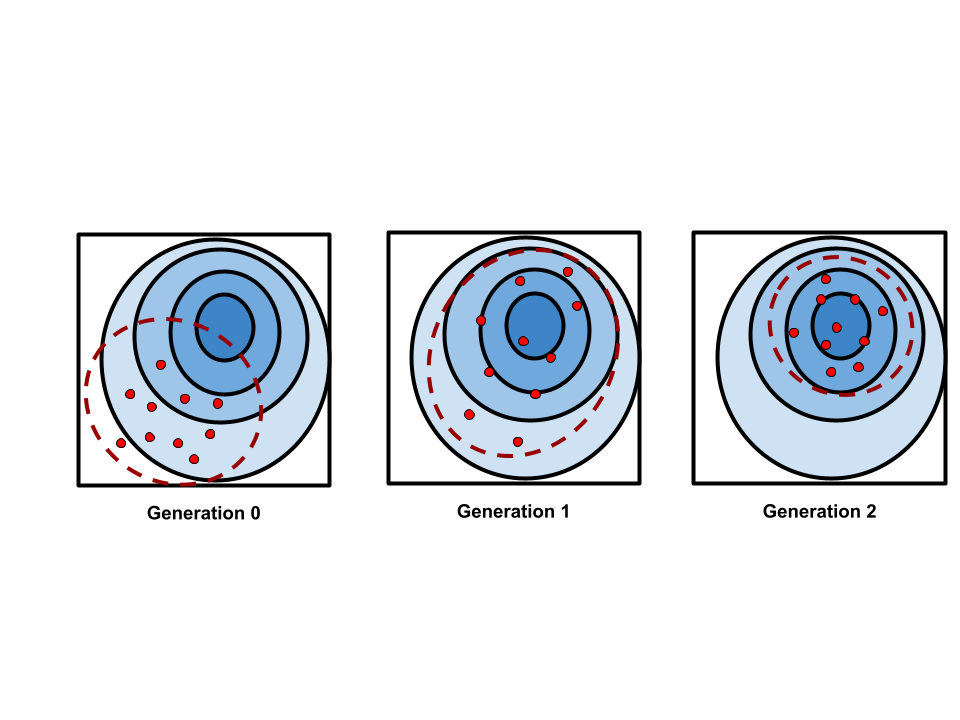}
    \caption{How CMA-ES produces fitter solutions by updating the search space given an unknown fitness landscape. More intense blue zones represent regions of higher fitness, red dots are generated genomes, and the dotted red ellipse represents the current search space.}
  \label{fig:drift}
\end{figure}

\subsection{Genome}

\begin{figure*}
    \centering
    \includegraphics[width=1.0\textwidth]{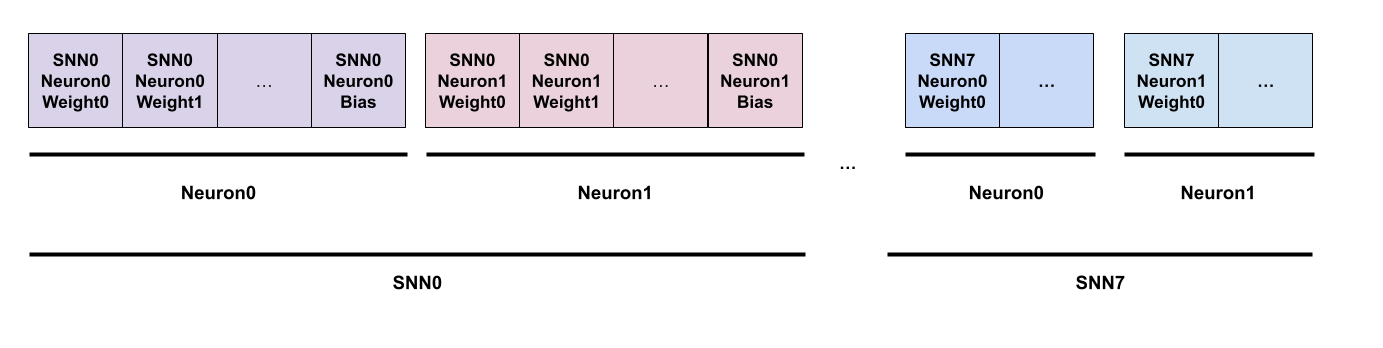}
    \caption{The structure of the flat genome}
  \label{fig:genome}
\end{figure*}

For our genome, we use a flat array of values that represent SNN weights and biases. Each actuator in the robot has its own independent SNN. Each SNN has two inputs, which are the distances from the voxel's center of mass to the top left and bottom right corners of the robot. From there, input values are sent to the network's hidden layer, which has two neurons. As each neuron in the hidden layer demands two separate weights for each incoming input layer neuron as well as a bias, this means the hidden layer account of six genome values. Finally, each SNN has one output node, which demands a further two weights from both neurons in the hidden layer. Factoring in the bias, this gives us a total of three necessary values for the output node, and thus nine for each SNN. With eight actuators in the robot, this gives us a total of 72 values for the genome. Figure \ref{fig:genome} demonstrates how these values are arranged in a flat array.

\section{SNN-Sim}

The SNN-Sim package is principally concerned with running the simulation and facilitating closed-loop robot control. To do this, robot telemetry is sampled, computed for use by the SNN, outputs from the SNN are received, and finally these values are set at actuator target lengths. This way, the robot can respond dynamically to its environment, as well as its current morphological state.

\subsection{Robot Instantiation and Control}

For our purposes, we make use of two Evogym methods for robot control.

\begin{itemize}
    \item \texttt{robot.sim.object\_pos\_at\_time(time: int, object\_name: str)}
    \begin{itemize}
        \item This returns the coordinates of all point masses in a given object in the for of a  \texttt{np.ndarray} of size $(2, n)$, where $n$ is the number of point masses in the object. The first subarray contains all $x$ coordinates, and the second subarray contains all corresponding $y$ coordinates.
    \end{itemize}
\item \texttt{sim.set\_action(robot\_name: str, action: ndarray)}
\begin{itemize}
    \item This method takes an "action array" of length $n$ where $n$ is the number of actuators in the robot. Thus, each array index corresponds to an actuators target length, and it will expand or contract to reach the desired size.
\end{itemize}
\end{itemize}

\begin{figure}
    \centering
    \includegraphics[width=1.0\columnwidth]{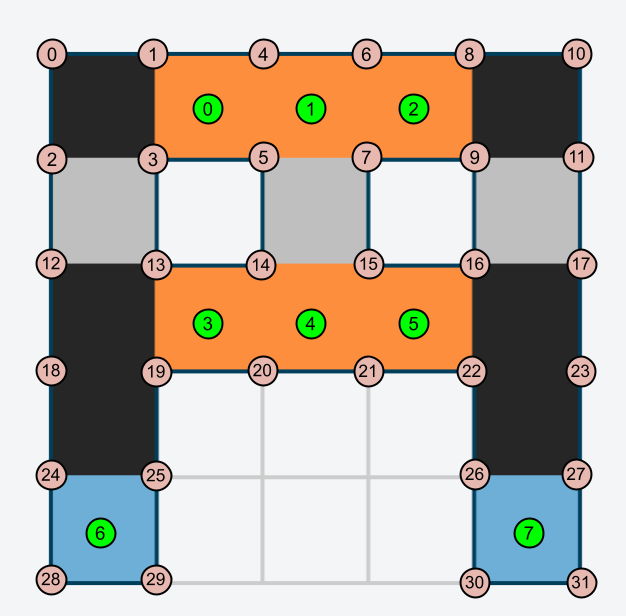}
    \caption{Action array indices (in green) and point mass indices (in red)}
  \label{fig:bestbot_numbering}
\end{figure}

However, it is important to recall that actuators and point masses are not equivalent: actuators are made of point masses. Each voxel, some of which are actuator voxels, has four point masses- one for each corner. In addition, adjacent voxels share point masses, so we must take care to parse point mass coordinates appropriately before sending them as inputs to the robot. Figure \ref{fig:bestbot_numbering} illustrates the index numbering, with the index of each point mass coordinate returned by \texttt{robot.sim.object\_pos\_at\_time()} highlighted in red, and the index of each actuator in the action array expected by \texttt{sim.set\_action()} highlighted in green. In general, Evogym iterates from left to right and then top to bottom, looking at each voxel in the robot. Actuator indices are assigned in this order. Point mass indices are also numbered in this order, with the top left point mass first, top left second, bottom left third, and bottom right last. However, if a point mass is shared with another voxel and has already been numbered, it is not renumbered. This leaves us with 8 actuators and 32 point masses for Ubot. 

A single simulation in Evogym consists of 1000 time steps, during each of which telemetry data can be extracted and new target lengths set for each voxel based on the SNN outputs. Since it takes approximately 10 time steps to allow a voxel to reach its full target length (whether that is a 0.6 contraction or 1.6 expansion), these inputs and lengths are sampled every 12 time steps, meaning there are 83 telemetry values and 83 calls to the SNN that produce movement behavior. Closed loop control is achieved by the following function calls:

\begin{enumerate}
    \item \texttt{raw\_pm\_pos = sim.object\_pos\_at\_time\\(sim.get\_time(), "robot")}
    \begin{enumerate}
        \item Store the raw point mass coordinates.
    \end{enumerate}
    \item \texttt{corner\_distances = np.array(morphology.get\_corner\_distances\\(raw\_pm\_pos))}
    \begin{enumerate}
        \item Uses our morphology class to convert point mass coordinates to distance from each actuator to the top left and bottom right corners of the robot using a point-mass to actuator mapping.
    \end{enumerate}
    \item \texttt{action = snn\_controller.get\_lengths\\(corner\_distances)}
    \begin{enumerate}
        \item Use corner distances as input the the SNNs. We use an SNN controller to instantiate the 8 different SNNs and manage SNN input.
    \end{enumerate}
    \item \texttt{sim.set\_action('robot', action)}
    \begin{enumerate}
        \item Set the target length of the robot's actuator's using the SNN output.
    \end{enumerate}
\end{enumerate}

\begin{figure}
    \centering
    \includegraphics[width=1.0\columnwidth]{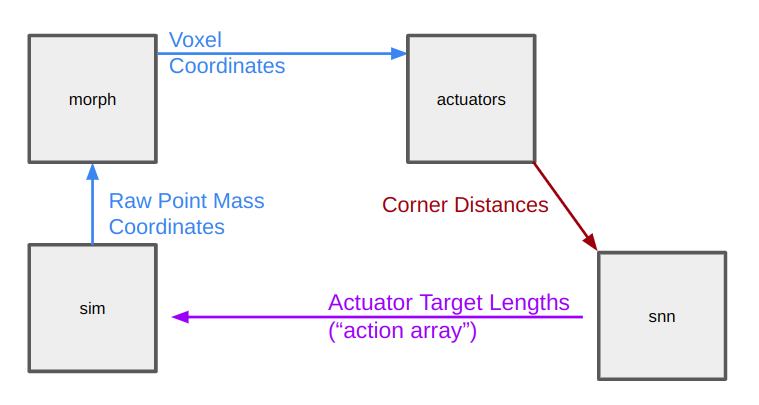}
    \caption{Closed loop robot control scheme}
  \label{fig:closed_loop}
\end{figure}

Thus, we achieve a closed-loop robot control system. Figure \ref{fig:closed_loop} illustrates the system in more detail. Supplemented with the entire framework diagram in Figure \ref{fig:complete_framework}, it is apparent how the morphology class, as our own internal representation of the robot, interacts with the SNN in order to ensure robot telemetry is processed effectively.

\subsection{Fitness Function}

A robot's fitness is assessed on its ability to move towards the end of a platform. Thus, the fitness function takes the average of all point mass $x$ coordinates before and after the simulation, and returns the different between the final position and the initial position. Thus, the fitness function can be defined as follows.

Let $X_i$ be the set of point mass $x$ positions at the start of the simulation, and let $X_f$ be the set of point mass positions at the end of the simulation.
$$f(X_i,X_f)= 100-\frac{\sum_{x \in X_f}x}{|X_f|}-\frac{\sum_{y \in X_i}y}{|X_i|}$$

Since the CMA-ES algorithm we are using is minimizing and our robot is tested on a platform of length 100, we subtract 100 from the difference of the final and initial positions. Thus, a theoretic perfect score would be 0.

\subsection{Spiking Neural Networks}
\subsection{Overview}

In this project, we developed a pipeline for evolving and executing soft robot locomotion using a Spiking Neural Network (SNN) controller. The SNN-based controller processes sensory inputs and generates actuation signals. This allows the robot to adapt and refine its movement over successive generations. Each actuator within the soft robot is controlled by a dedicated SNN instance, which operates independently but follows a unified architectural framework.

The evolutionary process optimizes robot behavior across generations by adjusting network parameters to enhance locomotion efficiency. Our approach utilizes the Covariance Matrix Adaptation Evolution Strategy (CMA-ES) to fine-tune the SNN parameters. While iteratively improving motor coordination and control.

\subsection{Spiking Neural Network Architecture}

The architecture consists of two main components: (1) the core SNN implementation, which defines the neuron dynamics, and (2) a separate module responsible for invoking the SNN instances and integrating them into the robot control framework.

\subsection{SpikyNode (Neuron)}

Each neuron in our model follows a leaky integration and fire (LIF) technique in which activation levels accumulate over successive time steps until the threshold is reached. A decay of 0.1 ensures that the activation level resets over time, preventing indefinite accumulation unless a spike occurs. This mechanism promotes a rhythmic spiking pattern, which is useful for periodic gaits such as jumping. The detailed properties of the neuron model are as follows:

\subsubsection{Neuron Dynamics}

\begin{itemize}
\item The neuron has an activation level that decays over time and changes according to the weighted sum of the inputs.
\item When the activation level exceeds a predefined threshold (bias), the neuron fires, releasing a value of 1.0. If the threshold is not reached, it outputs a value of 0.0.
\item The output is scaled to control the behavior of the soft robot voxel, where the values range from \textbf{0.6 to 1.6}. A value of 0.6 contracts the voxel, while 1.6 expands it.
\item A \textbf{fire log} tracks the firing history and it is implemented as a circular buffer for efficiency.
\end{itemize}

\subsection{Network Structure}

The SNN is structured as a \textbf{two-layer network}, consisting of:

\begin{itemize}
\item \textbf{Hidden Layer:} Contains two neurons that process input information and modulate the propagation of activation.
\item \textbf{Output Layer:} Contains one neuron that directly governs actuator behavior, generating spike-based control signals.
\item \textbf{Per-Actuator Networks:} Each actuator within the soft robot is assigned an independent SNN instance, allowing localized control and modular learning.
\item \textbf{Flexible Architecture:} The input size, hidden layer size, and output size are adjustable, allowing for adaptability across different robotic configurations.
\end{itemize}

\subsection{Controller and Learning Process}

The \textbf{SNN controller} orchestrates the behavior of the soft robot by:

\begin{enumerate}
\item \textbf{Creating a dedicated SNN instance} for each actuator.
\item \textbf{Processing sensory inputs}, specifically the distances between selected voxels and two reference corners of the robot.
\item \textbf{Generating spike outputs} based on the accumulated membrane potential and predefined thresholds.
\item \textbf{Applying CMA-ES} to evolve the network parameters over generations.
\end{enumerate}

The evolutionary process optimizes the network by:

\begin{itemize}
\item Encoding of the SNN parameters (weights and biases) as a flat vector.
\item Using CMA-ES to adjust these parameters based on fitness evaluations.
\item Reshaping the evolved parameter vector into a structured format for each SNN instance.
\item Iterating this process over multiple generations to enhance locomotion performance.
\end{itemize}

This approach enables the soft robot to autonomously refine its movement strategy while leveraging SNNs for adaptive and dynamic control. Over time, the network evolves its structured, rhythmic actuation patterns that improve locomotion efficiency. All while demonstrating the potential of SNN-based learning in robotic applications.
\section{Results}

\subsection{Initial Results}

The initial results obtained using the Spiking Neural Network (SNN) framework leveraged several mechanisms that produced relatively fit examples but fell short of the primary objective: exhibiting morphological communication. 

In the initial stages, many robots exploited a simple yet highly effective strategy - falling over to move in the x-plane and therefore gain fitness. This behavior provided an easy pathway to increasing fitness, making it very difficult for the evolutionary algorithm to develop more complex locomotion. Since many of the fittest individuals in the population-based search had adopted this behavior, the algorithm struggled to escape the local optima, instead making subtle adjustments to the way the robots fell and which voxels triggered the falling behavior to achieve incremental advances.

\begin{figure}[ht]
    \centering
    \includegraphics[width=0.2\textwidth]{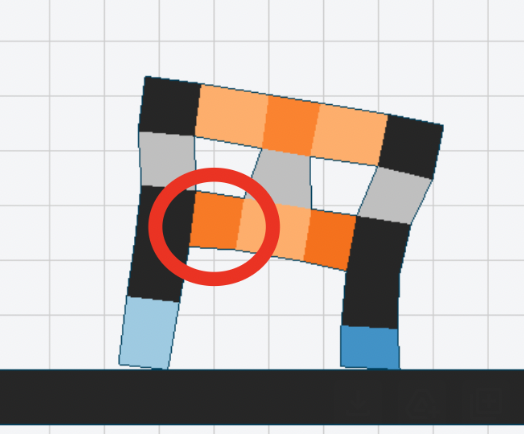}
    \caption{Robot exploiting a single voxel for locomotion}
    \label{fig:Single_voxel_exploit}
\end{figure}

After mitigating this issue by tracking the y-coordinates of the robot's legs and center of mass and penalizing robots that had fallen over, another interesting behavior developed. Robots began to exploit a single voxel to drive their movement and achieved high fitness. The example in Figure \ref{fig:Single_voxel_exploit} used a horizontal actuator paired with an expanded left leg and contracted right leg to push itself along the platform. Again, these discovered genomes were local optima with only one set of SNN weights adequately evolved.

\subsection{Refinement}

The absence of complex locomotion in the initial examples can be attributed to the lack of spikes occurring within each voxel's SNN. A spike, which increases the target length of an actuator causing it to expand, resets the SNN's membrane potential, causing a likely contraction at the next sample. Increasing the likelihood of spikes by tuning the SNN and scaling input lengths by their initial lengths (target lengths of 1 for every actuator) caused more complex behaviors to develop since multiple actuators were inclined to change target lengths at once. This had the double benefit of creating more pronounced movement within the robot, which decreased input sensitivity by amplifying the changes in corner distances.

\begin{figure}[ht]
    \centering
    \includegraphics[width=0.4\textwidth]{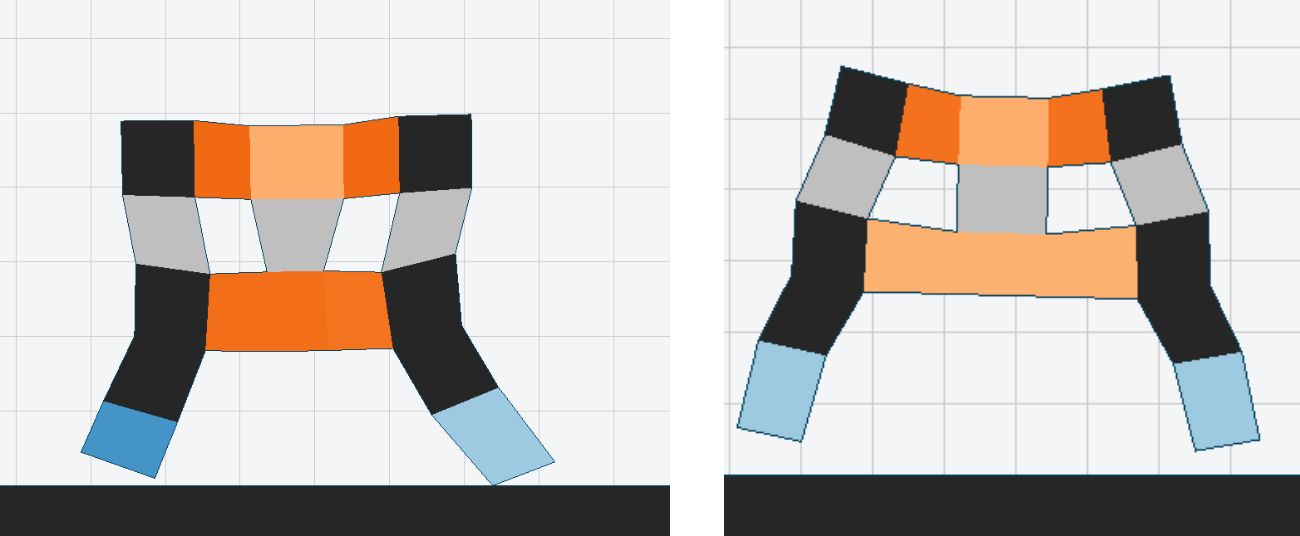}
    \caption{Robot using multiple actuators at once}
    \label{fig:Morpho_Comm_1}
\end{figure}

As a result, more complex behaviors began to develop. Figure \ref{fig:Morpho_Comm_1} depicts a robot that has evolved to use its three middle actuators at the same time. In the first image, the three are contracted, and in the second, they are expanded to create a jumping gait. This was the first robot to develop that was able to reach the end of the platform, taking 64 generations to do so.

\begin{figure}[ht]
    \centering
    \includegraphics[width=0.48\textwidth]{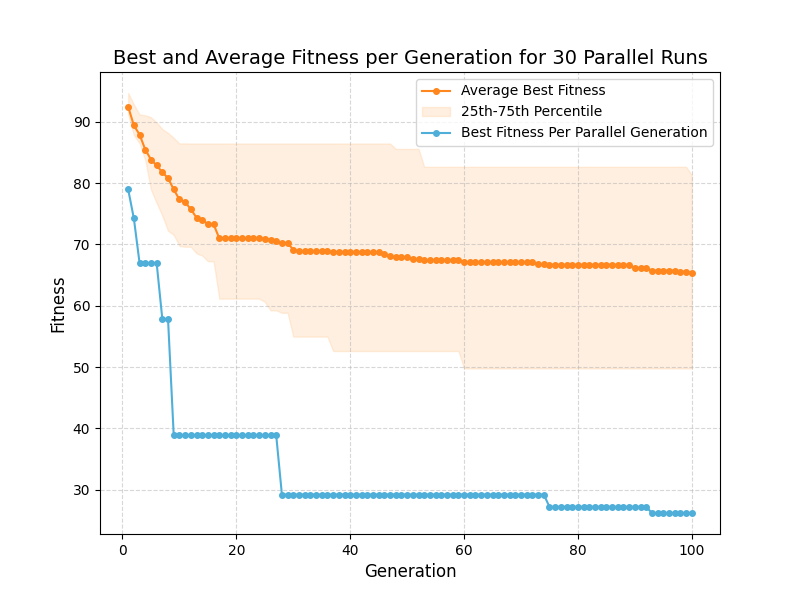}
    \caption{Average fitness over generations}
    \label{fig:Fitness_Gens_64}
\end{figure}

With any evolutionary algorithm, there exists a trade-off between performance and evaluation time. Figure \ref{fig:Fitness_Gens_64} provides a breakdown of fitness improvements over generations for the robot in Figure \ref{fig:Morpho_Comm_1}.

This experiment was completed with 30 parallel populations to cut down processing time and increase genetic diversity without significantly increasing computational cost. The graph shows a clear decreasing trend from the initial fitness of 100 towards the best fitness achieved of 27 using the 0 to 100 fitness scale. The returns diminish over time, with a fitness below 30 being reached in the 28th generation, and an improvement of only 2 in the subsequent 62 gens. 

\subsection{Morphological Communication}

As stated, individual actuators have no explicit knowledge of the control methods for other actuators within the robot. The only driving behavior for a voxel's expansions and contractions is due to telemetry data, and therefore the only way the voxels can communicate with one another is tacitly. When a voxel expands or contracts it changes the corner distances for every other voxel within the robot. These changes are minor in isolation, considering a target length can only vary between 0.6 and 1.6, but when voxels begin to work in tandem a self-confirming loop is initiated whereby large changes in telemetry are absorbed by the SNNs and drive even more complex behavior.

In most highly evolved examples the six horizontal actuators evolved to work in tandem, with equivalent target values for each step of the simulation. However, the behavior of the two vertical actuators is perhaps more interesting. Despite the inputs to these actuators being simple target lengths, in many highly evolved robots they do not follow the simple, expand, contract, expand pattern, instead developing an offset or triggering only a couple of times across a 10,000 step simulation.

\begin{figure}[ht]
    \centering
    \includegraphics[width=0.45\textwidth]{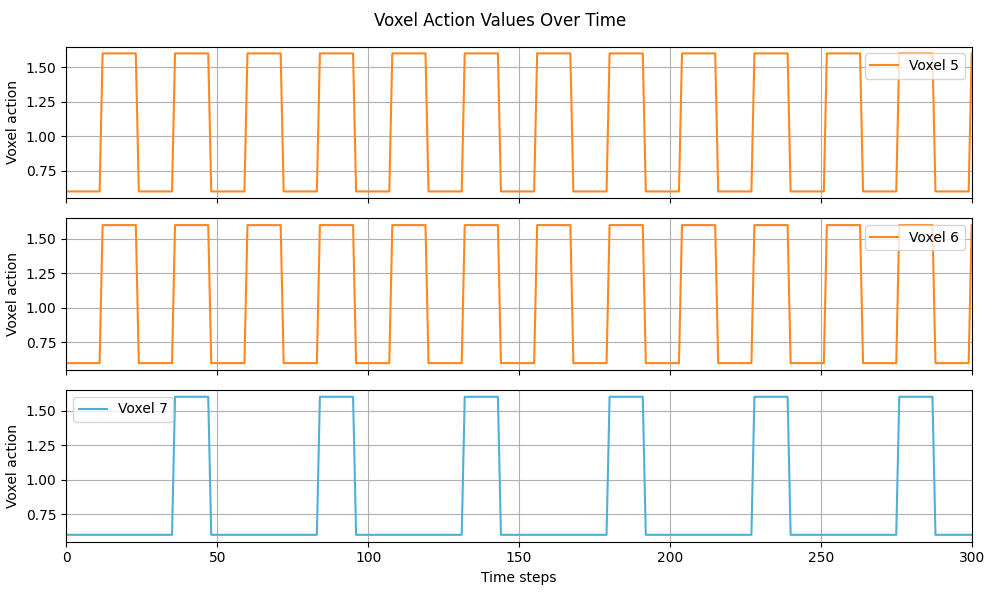}
    \caption{The action values of three voxels in an evolved robot}
    \label{fig:voxel_lengths}
\end{figure}

For a robot that developed identifiable morphological communication between the horizontal actuators, the bottom left vertical actuator in Figure \ref{fig:Morpho_Comm_1} exhibits a more complicated behavior. The plot is high after a spike in the neural network which sets the actuator's target length to 1.6, and low when the network does not spike and the length is set to 0.6. The voxel's "base state" is contracted and it only actuates when the left leg is in contact with the ground and in a way that is beneficial to the robot's gait. In fact, this actuation pattern aligns with a specific orientation of the robot which allows the left leg to propel it forward due to the angle of the actuator. This pattern is due to the membrane potential level of the actuator's SNN. It takes two inputs for the level to accumulate above the bias after which the target length is set to 1.6, this pattern repeats after the distance is reset.

\section{Discussion}

There is a significant difference between the minimization time and generations from one run to another. Parallelism is often needed to introduce enough genetic diversity at the start of the evolutionary process. Since sigma decreases over time, CMA-ES may stagnate at a local optima, hence why the high percentile does not appear to improve significantly. Failure to initiate parallel CMA-ES optimizers with different populations runs the risk of finding one of these populations that does not significantly improve. In the worst case fitness does not develop beyond 90 (using the 0-100 fitness function) and the best robot produced does not exhibit complex locomotion behavior or morphological communication.

One optimization we tried was implementing Spike Timing Dependent Plasticity, where the weights and biases were increased by 5\% if the nodes were spiking too many times, and decreased by 5\% if the nodes were not spiking enough. This improvement resulted in some voxels being activated more than before and the robot as a whole exhibiting slightly more complex movements but it was not stable enough to actively use throughout the project. A better method to increase efficiency could be to prune branches that do not exhibit fitness less than 50 after n generations since our experiments suggest the minimization becomes more complex at this stage and therefore robots stuck here are lost causes. This strategy could be implemented to copy the population of a fit evolution thread and create a new parallel CMA-ES thread.

\section{Future Work}

While this work establishes a basis for conducting experiments in morphological communication in Evogym, there are many avenues of future research which could be investigated. More could be done to optimize robot fitness by adjusting CMA-ES hyper-parameters. By further investigating different values for the initial mean genome, sigma, and fitness progression over generations, we may be able to achieve fitter robots faster and accelerate the time needed to produce viable robot control structures.

Future work may investigate alternative ways for robot actuators to fire based on SNN outputs. Our final implementation sets robot actuators to their fully extended position when SNN duty cycle equals 1; however, other approaches could be used, such as forgoing the duty cycle and allowing for SNN spikes to trigger motor actuation directly. Future work may look to investigate how quickly neuron level accumulates given different SNN weights and biases.

We experimented with throttling SNN inputs to once every 12 Evogym time steps to prevent over-spiking and account for the time required for voxels to reach their target lengths. Additionally, we normalized SNN inputs. While some robots developed more complex behaviors, where voxels expanded and contracted in a nuanced pattern, we did not observe evolution toward full expansion or contraction; instead, robots tended to vibrate. Future work could further refine the SNN or adjust input values to enhance the network’s sensitivity to telemetry changes, enabling the SNN to "discover" the 12-timestep pattern autonomously.

Other areas of inquiry include the development alternative robot designs, perhaps suited for tasks such as jumping or moving an object, or producing an objective metric to assess the locomotive potential of a robot design. Other robot control methodologies could also be tested, including alternative SNN inputs which may offer more promising avenues for developing morphological communication.

\section{Acknowledgements and Contributions}

\begin{enumerate}
    \item We are very thankful to Jagdeep Singh Bhatia, Holly Jackson, Yunsheng Tian, Jie Xu, and Wojciech Matusik for their work in developing the EvoGym simulator, as well as the extensive documentation that aided us in creaing the experimental framework featured in this paper.
    \item The Spiking Neural Network (SNN) team, consisting of Abhay Kashyap, Atharv Tekurkar, Jonathan Fischman, Luodi Wang, and Miguel Garduno, was responsible for the research and implementation of the SNN based on original C code written by John Rieffel. They also implemented an SNN wrapper class for the rest of the group to access and use the SNNs along with some processing and utility methods.
    \item The robot morphology used in experiments and in providing the paper's results was developed by the Morphology team. Matthew Meek, Guy Tallent, and Viet-Dung "Darren" Nguyen used evolutionary algorithms featured in the Evogym \cite{EvoGym} package to develop the robot's morphology. They primarily worked to integrate Evogym's simulation framework into the broader project framework with CMA-ES and the SNNs, implement code for capturing the necessary telemetry, and help the other teams navigate the Evogym package.
    \item The experimental framework team, consisting of Thomas Breimer and James Gaskell, developed the CMA-ES optimization pipeline including integration of the morphology and SNN, parameter optimization, and application of telemetry data from Evogym's \cite{EvoGym} simulation modules. They also developed auxiliary programs for data capture, visualization, and experiment isolation which facilitated the analysis and presentation of the results in this paper.
\end{enumerate}

\footnotesize
\bibliographystyle{apalike}
\bibliography{bibliography}

\end{document}